\documentclass[letterpaper, 10 pt, journal, twoside]{ieeetran}
\usepackage{amsmath,amsfonts}
\usepackage{algorithmic}
\usepackage{algorithm}
\usepackage{array}
\usepackage[caption=false,font=normalsize,labelfont=sf,textfont=sf]{subfig}
\usepackage{textcomp}
\usepackage{stfloats}
\usepackage{url}
\usepackage{verbatim}
\usepackage{graphicx}
\usepackage{cite}
\def\ourmethod{SensorPerch}
\usepackage{times}
\usepackage{graphics} 
\usepackage{epsfig} 
\usepackage{times} 
\usepackage{amsmath} 
\usepackage{amssymb}  
\usepackage{tcolorbox}
\usepackage{algorithm}
\usepackage{amsmath, amssymb}
\usepackage{caption}
\usepackage{multirow}
\usepackage{graphicx}
\usepackage{bbm}
\usepackage{booktabs}

\usepackage{hyperref}
\hypersetup{colorlinks=true, urlcolor=blue, linkcolor=black, citecolor=black}
\usepackage{xcolor}

\begin{document}

\title{SensorPerch: Sense Wherever and Whenever it Matters}

\author{
Zhanxin Wu$^{1,*}$,
Ruofei Tong$^{1,*}$,
and Tapomayukh Bhattacharjee$^{1}$

\thanks{This work was partly funded by National Science Foundation IIS \#2132846, and CAREER \#2238792. This research was also funded, in part, by the Advanced Research Projects Agency for Health (ARPA-H) Agreement No. 140D042590012. The views and conclusions contained in this document are those of the authors and should not be interpreted as representing the official policies, either expressed or implied, of the U.S. Government.} 
\thanks{$^{1}$Department of Computer Science, Cornell University, Ithaca, New York, United States. 
{\tt\small zw754@cornell.edu}, {\tt\small rt455@cornell.edu}, {\tt\small tapomayukh@cornell.edu}.
}
\thanks{$^{*}$Equal Contribution.}
}

\markboth{}%
{Wu \MakeLowercase{\textit{et al.}}: SensorPerch: Sense Wherever and Whenever it Matters}


\IEEEaftertitletext{%
    \vspace{-6pt}
    \begin{center}
        \includegraphics[width=0.9\textwidth]{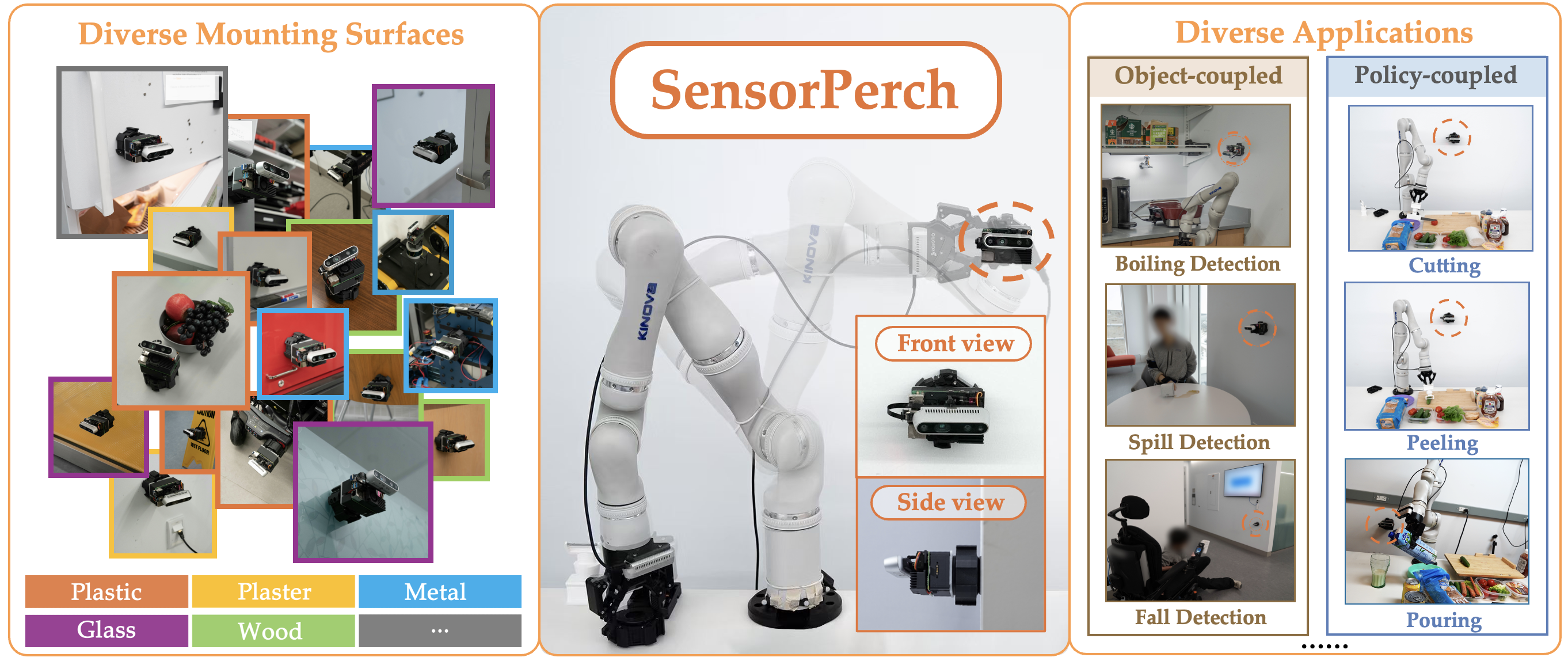}
        \captionsetup{hypcap=false}
        \captionof{figure}{%
            \textbf{\ourmethod{}} decouples sensing from both
            robot embodiment and the environment by treating
            sensors as independent physical entities that the
            robot can autonomously detach and reattach
            in the environment.
        }
        \label{fig:teaser}
    \end{center}
}

\maketitle

\begin{abstract}
Existing robotic perception is constrained by sensors that are either robot-mounted or permanently fixed in the environment, locking perception to a limited set of viewpoints. Yet as robots perform increasingly diverse tasks, the most informative viewpoint shifts from one task to the next—often somewhere onboard sensor and static infrastructure cannot readily satisfy. To address this gap, we propose \ourmethod{}, a novel realization of active perception that decouples sensing from both the robot embodiment and the environment by treating sensors as independent physical entities that the robot can autonomously detach and reattach within the environment. \ourmethod{} presents one realization of this paradigm: a lightweight, wireless, reconfigurable sensor platform that can perch on diverse surfaces, paired with a viewpoint-selection framework that determines task-optimal sensor placements. Together, these enable robots to construct task-relevant viewpoints on demand, independent of the robot's current position and available fixed infrastructure. We demonstrate the paradigm on two task classes: (i) object-coupled perception, where \ourmethod{} enables persistent object-state detection beyond the robot's current position, achieving successful event detection even when the robot is not nearby; and (ii) policy-coupled perception, where \ourmethod{} allows robots to construct diverse, policy-specific viewpoints for various policies, achieving success rates comparable to those obtained using oracle viewpoints. 
Website: \href{https://emprise.cs.cornell.edu/sensorperch/}{https://emprise.cs.cornell.edu/sensorperch/}.
\end{abstract}


\section{Introduction}

\IEEEPARstart{C}{onsider} a robot preparing breakfast in a kitchen and then stepping away to fold laundry while waiting for a pot to boil. The moment it leaves, information outside its field of view, such as the pot beginning to boil over, becomes inaccessible even though it remains critical. As robots take on increasingly diverse tasks, this problem compounds, since each task demands its own viewpoint. Preparing breakfast may benefit from an angled view of the kitchen counter, while folding clothes may require a top-down view of the laundry table. As the robot transitions between tasks and locations, the set of task-relevant viewpoints shifts accordingly~\cite{intelligence2025pi05visionlanguageactionmodelopenworld}. Yet existing robotic perception systems effectively lock sensing to a limited set of viewpoints: sensors are either rigidly mounted on the robot embodiment or permanently fixed in environmental infrastructure. Perception is therefore constrained to viewpoints the robot can physically occupy at a given moment, or those covered by pre-installed fixed infrastructure. This tight coupling between perception, robot embodiment, and infrastructure fundamentally limits a robot's ability to perceive what each task requires.

Robotic perception has traditionally followed two paradigms: passive perception and active perception. Passive perception from fixed viewpoints is vulnerable to occlusion as the environment changes~\cite{slam++2013, Girshick2013RichFH}. Active perception extends this paradigm by moving onboard sensors during task execution, for instance by repositioning a wrist-mounted camera or articulating a robotic neck or head~\cite{xiong2025via, 2019Humanoidhonda}. While these approaches increase viewpoint flexibility, they share a common but often implicit assumption: sensors are constrained by the robot's physical embodiment and the pre-installed infrastructure in the environment. Consequently, obtaining information outside the current field of view requires interrupting task execution to relocate the robot or manually setting up additional viewpoints.

 \begin{figure*}[t]
 \vspace{2mm}
     \centering
     \includegraphics[width=\linewidth]{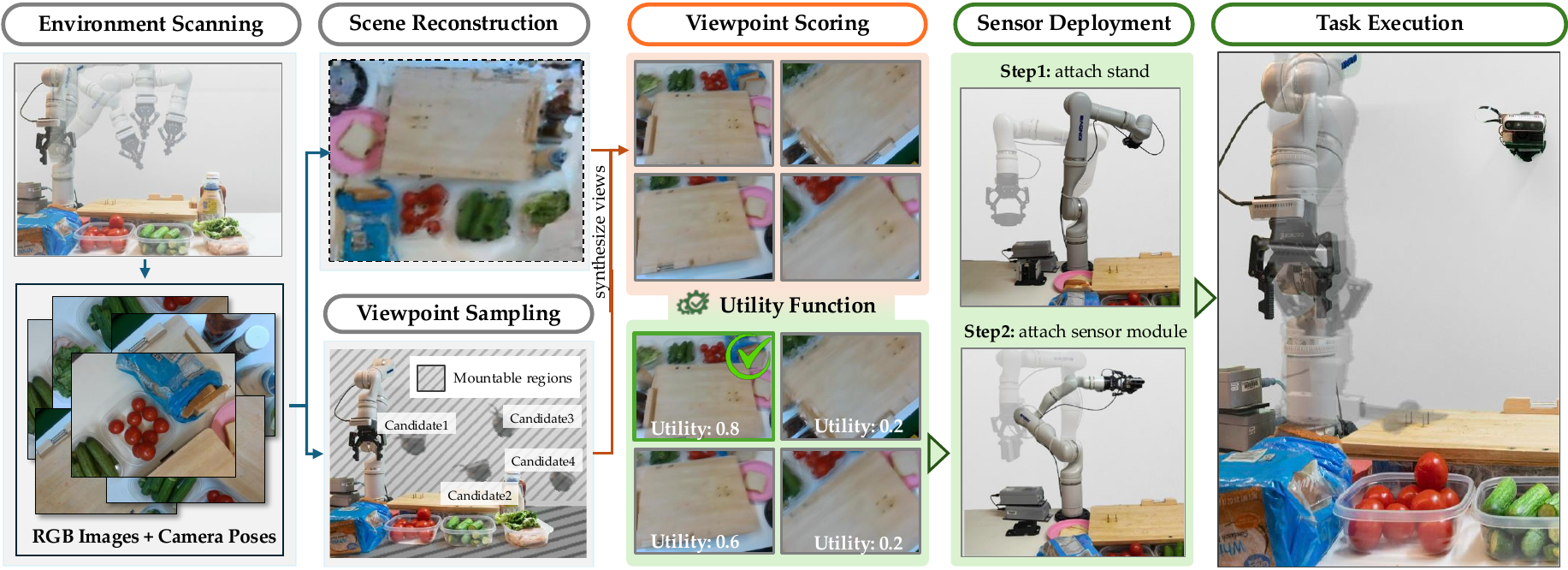}
    \caption{
    \textbf{Overview.} We first scan the environment and reconstruct a simulated scene. We then identify mountable regions and sample candidate viewpoints. For each candidate, we synthesize novel views and evaluate them using a task-conditioned utility function. The robot then attaches the \ourmethod{} platform at the highest-utility viewpoint to support downstream tasks.
    }
     \label{fig:framework}
     \vspace{-12pt}
 \end{figure*}

Our key insight is that task-adaptive perception hinges on pairing a hardware capability, namely sensors the robot can autonomously detach and reattach on demand, with a framework that decides where to reposition them. We propose a novel realization of active perception, \ourmethod{} (Fig~\ref{fig:framework}), that treats sensors as independent physical entities that the robot can autonomously detach and reattach within the environment, so that sensors can be placed wherever a task needs them. Sensors are neither permanently attached to the robot’s kinematic structure nor installed as fixed infrastructure; instead, they temporarily perch in the environment. Our paradigm is agnostic to any particular sensor, attachment mechanism, or placement strategy. We present one instantiation: a lightweight, wireless, reconfigurable sensor platform that can physically perch on environment surfaces, paired with a framework that decides where each task's sensing should occur.
Our sensor platform provides two degrees of freedom for viewpoint adjustment and uses a compact vacuum-based attachment mechanism that enables reliable attachment on diverse surfaces. The framework samples candidate viewpoints at which the robot can feasibly deploy the platform and selects the one that best serves the task. This closed-loop reasoning enables the robot to deploy the sensor at the most informative viewpoint.

We evaluate \ourmethod{} with real-robot experiments across two task classes spanning eight tasks: object-coupled perception, which enables persistent object-state estimation beyond the robot's current position, and policy-coupled perception, which autonomously places sensors at policy-required viewpoints for downstream inference. Across these experiments, \ourmethod{} successfully detects object states beyond the robot's immediate field of view and achieves policy-execution success rates comparable to oracle third-person viewpoints, demonstrating that perception can be treated as a reconfigurable resource rather than a capability fixed by embodiment or pre-installed infrastructure.

\section{Related Work}
\label{sec:RelatedWork}

\textbf{Active perception in robotics.} Existing work in active perception has explored how a robot can control its sensing configuration during task execution~\cite{Aloimonos2004ActiveVision, tsotsos1995modeling}. {\color{black}Rather than passively accepting observations, the robot actively seeks informative viewpoints that reduce uncertainty or improve task-relevant perception~\cite{cheng2018reinforcement, shang2023active, jayaraman2018learning, strong2024nextbestsense}.} 
Many systems incorporate hardware mechanisms to support active sensing, including articulated camera mounts~\cite{1999_NBV_hardware}, anthropomorphic head and eye movements~\cite{ 2019Humanoidhonda}, and dual-arm setups in which one arm repositions a camera to observe manipulation~\cite{kappler2018real, xiong2025via, lenz2023bimanual}. 
However, in these onboard sensing paradigms, sensing remains fundamentally constrained by the robot's embodiment: sensors are permanently attached to the robot's kinematic structure, and such perceptual access is lost once the robot moves away. 
In contrast, \ourmethod{} enables sensing that is physically decoupled from embodiment and environment. 
This allows the robot to proactively detach and reattach sensors to various positions as needed.

\textbf{Multi-view perception in robotics.} Prior work has leveraged multiple viewpoints to support task execution~\cite{hsu2022vision, jangir2022look, arnold2020cooperative}. A common strategy is to surround the workspace with multiple static cameras, providing broad coverage and reducing occlusions~\cite{chen2023predicting, andrychowicz2020learning}. This idea has been further extended to pre-instrumented smart environments~\cite{Pyo2015ServiceRS}, where sensors are embedded in infrastructure, and to multi-robot systems in which robots share perceptual information to form a distributed sensing network~\cite{hudson2022heterogeneous, zhou2023racer}. Robots have also deployed sensors into the environment, either as network nodes~\cite{corke2004autonomous} or by installing them on structures with aerial robots~\cite{hamaza2019sensor, farinha2020unmanned}. While these approaches improve environment coverage, they rely on fixed infrastructure, coordinated multi-agent setups, or placements chosen for coverage and connectivity rather than task needs. Such systems require careful calibration and manual reconfiguration as tasks or environments change, limiting their flexibility in everyday settings. In contrast, \ourmethod{} enables a single robot to reposition its own sensors on demand, constructing task-adaptive viewpoints without relying on permanent infrastructure or multi-robot coordination.

\textbf{Viewpoint optimization for robot manipulation.} Prior work on viewpoint selection has focused on geometric and perceptual criteria, such as maximizing coverage, reducing pose uncertainty, or improving reconstruction quality, typically casting view planning as an optimization problem to enhance visual observability~\cite{InteractivePerception2017Jeannette, NBV_2016, zeng2020viewsurvey, pan2022activenerf, chen2026cover, goli2024bayes}. More recent approaches incorporate task context by selecting viewpoints that are informative for evaluating action outcomes, often in static or dynamic simulated environments~\cite{ning2025prompting, li2025actloclearninglocalizeactive, shen2023distilled, kerr2024robot}. In these methods, viewpoint selection primarily serves as an information-gathering mechanism to guide planning. In contrast, we treat viewpoint selection as a task-conditioned optimization problem that determines not only which views are informative, but also which should be physically constructed in the real world. By coupling viewpoint selection with reconfigurable sensing modules, we enable flexible third-person perception for downstream tasks.

\begin{figure*}[ht]
\vspace{2mm}
    \centering
    \includegraphics[width=\textwidth]{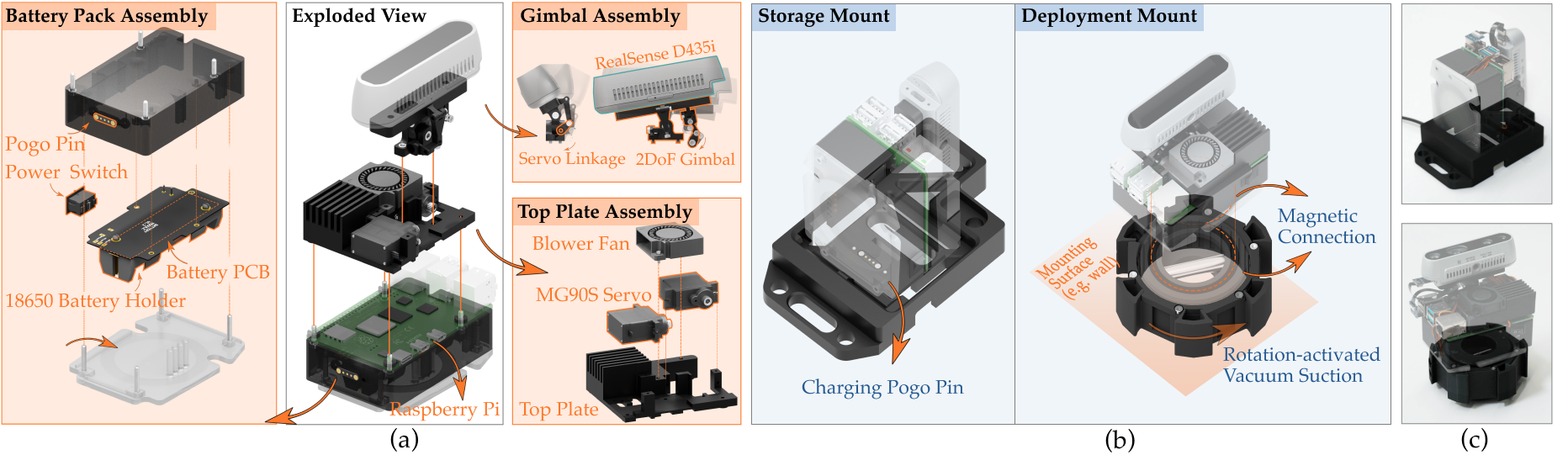}
    \caption{
    \textbf{Hardware overview of \ourmethod{}.}
    (a) Exploded views illustrate the modular architecture, including the battery pack, computation module, and servo-driven gimbal with a camera. 
    (b) Interchangeable mounting strategies, including a vertical dock with continuous power and a standalone magnetic suction stand.
    (c) Real-world hardware.
    }
    \label{fig:hardwaremerge}
    \vspace{-12pt}
\end{figure*}

\section{Problem Formulation}
\label{sec:ProblemFormulation}
We consider the problem of deploying sensors to support diverse robot manipulation tasks. The robot operates in a partially known environment $\mathcal{E}$ (e.g., a cluttered home) with $N$ sensor modules. Let $s_i$ denote the $i$-th sensor module; each has known intrinsic parameters and a known field of view. We denote its viewpoint by a 6-DoF pose $v_i \in SE(3)$ with respect to the world frame. A sensor placement configuration is the set of chosen viewpoints $V = \{v_1, \ldots, v_N\}$. A manipulation task $\tau$ can be decomposed into a sequence of subtasks $\tau = (\tau_1, \ldots, \tau_K)$, each imposing distinct perceptual requirements. For example, meal preparation may include boiling water, pouring, peeling, and cutting, each of which may be best observed from a different viewpoint.
To capture this, we define a task-conditioned perceptual utility
\[
U(V, \mathcal{E}, \tau_k),
\]
which evaluates how well a configuration $V$ satisfies the perceptual requirements of subtask $\tau_k$ in environment $\mathcal{E}$. Higher values of $U$ indicate better perceptual support. For each subtask, the goal is to select the configuration that maximizes this utility: \[ V_k^\star = \arg\max_{V} \; U(V, \mathcal{E}, \tau_k). \] This formulation captures the core challenge of task-adaptive sensor placement: because different tasks impose differing perceptual requirements, no single static configuration serves the whole task. Therefore, the robot must instead automatically reconfigure its sensors to maximize perceptual support.



\section{\ourmethod{}: How to perch}
\label{sec:hardware}

To enable the robot to autonomously attach and detach its sensors, the hardware must meet two requirements. First, each sensor platform must operate independently, sustaining itself throughout task execution and streaming observations with low latency. Second, it must be physically attachable and detachable at surfaces throughout the environment. We therefore introduce a modular sensor platform that decouples sensing from both the robot's embodiment and fixed infrastructure. To operate independently, each platform (Fig.~\ref{fig:hardwaremerge}a) integrates an onboard computation unit, a two-degree-of-freedom motorized gimbal with active cooling, and a battery pack. To make attachment and detachment repeatable over long-horizon operation, we further develop an ecosystem of charging stations mounted on the robot's mobile base, together with deployment mounting strategies (Fig.~\ref{fig:hardwaremerge}b). 
We detail each component below.

\subsection{Computation Platform} 
To support streaming diverse sensing modalities with low latency, the sensor module requires onboard computation for heterogeneous sensor interfaces, local processing, and wireless communication while remaining compact and energy-efficient. Therefore, we use a Raspberry Pi 4B (Fig.~\ref{fig:hardwaremerge}a) as the onboard compute platform, providing support for interfaces and communication protocols including I2C, SPI, UART, PWM, GPIO, and high-speed USB devices.


The Raspberry Pi interfaces with the sensor and gimbal servos, encodes the sensor stream using a GStreamer pipeline, and transmits it via low-latency User Datagram Protocol (UDP) over a shared IP network to a remote workstation. A lightweight receiver node on the workstation decodes the stream and republishes it as a standard ROS topic, enabling seamless integration with existing perception pipelines. In parallel, high-level servo commands are sent to the Pi over UDP, where a PWM controller drives the gimbal orientation. We provide the interface as a lightweight ROS package supporting both ROS1 and ROS2.

\subsection{Gimbal Assembly and Radiator}


The gimbal expands the field-of-view of onboard sensors by providing pitch and roll adjustments with adjustment ranges of $40^\circ$ in roll and $30^\circ$ in pitch. The top plate embeds an active radiator to prevent thermal throttling when the Raspberry Pi operates at high power. To ensure sufficient thermal conductivity, the plate is machined from metal. The entire assembly fits within the Raspberry Pi footprint to minimize module size.



\subsection{Battery and Charging Assembly}


To power the platform, we design a compact battery module using a single 18650 lithium-ion cell (Fig.~\ref{fig:hardwaremerge}, Battery Pack Assembly). This form factor provides a nominal capacity of approximately 3300~mAh while remaining compact enough to fit within the battery housing. The cell supports up to 10~A continuous discharge, delivering $\sim$30~W to the platform.



\subsection{Mounting Mechanism}
We present two mounting mechanisms in Fig.~\ref{fig:hardwaremerge}b: a \textit{storage mount} and a \textit{deployment mount}. We install the \textit{storage mount} on the robot's mobile base, allowing the robot to carry and recharge the platform. A male pogo-pin connector on the storage mount mates with a female connector on the sensor platform, enabling automatic charging whenever the platform is docked.
The \textit{deployment mount} attaches the sensor platform to surfaces in the environment. It uses a rotation-activated vacuum stand: the robot rotates the suction stand base in one direction to create a vacuum seal against the surface, and rotates it in the opposite direction to release the seal. The suction stand and sensor platform connect through a magnetic interface: placing the platform against the stand automatically seats it, and the robot removes it by grasping and applying a force that overcomes the magnetic attraction. This enables reliable detachment and attachment.


\subsection{Putting Everything Together}
The \ourmethod{} platform weighs 201.4g and has external dimensions of 99 mm × 66 mm × 67 mm. The total hardware cost is approximately \$94, including a Raspberry Pi 4B (\$60), two MG90S servos (\$10), a blower fan (\$4), a PCB (\$5), an 18650 battery cell (\$6), a pogo-pin connector (\$4), and miscellaneous hardware and filament (\$5).

\section{\ourmethod{}: Where to Perch}
The software stack of \ourmethod{} determines \emph{where} to place sensors to best support a task. 
In principle, the robot could evaluate viewpoints by physically placing the sensor at each candidate and capturing real observations; in practice, exhaustively trying every candidate is prohibitively slow. We therefore present a realization that decouples viewpoint evaluation from physical placement: we reconstruct the scene once, evaluate all candidate viewpoints in simulation, and commit only the highest-utility placement to the real world. This is one instantiation of the framework; other choices, such as alternative scene reconstruction methods, are also compatible with the framework. We detail each component below.

\subsection{Scene Reconstruction}
We maintain two complementary scene representations.
We use Kimera~\cite{Rosinol20icra-Kimera} to construct a metric-semantic map used for geometric reasoning and determining attachment feasibility, while a radiance-field model $\mathcal{R}$ (SVRaster~\cite{svraster}) learned from captured RGB images provides view-dependent appearance for novel view synthesis.
The radiance-field representation enables real-time rendering of observations from candidate viewpoints within the reconstructed scene (Sec.~\ref{sec:viewpoint_sampling_and_synthesize}) for subsequent viewpoint evaluation. 
This separation allows us to reason about \emph{where} sensors can be physically deployed while evaluating \emph{what} each viewpoint is expected to observe.

Constructing the metric-semantic map requires RGB-D observations, per-pixel semantic labels derived from RGB images, and camera poses expressed in a metric coordinate frame. Learning the environment’s radiance-field representation $\mathcal{R}$ requires RGB images and corresponding camera poses. To support both representations across different deployment scenarios, our system provides two data acquisition modes: robot-assisted scanning and human-assisted scanning.

\textbf{Robot-assisted scanning.}
The robot performs an exploratory scan using a wrist-mounted RGB-D camera while recording calibrated camera poses from forward kinematics. This produces metrically scaled images with known extrinsics in the robot frame.

\textbf{Human-assisted scanning.}
A human operator uses either the \ourmethod{} platform or a smartphone to capture RGB images of the environment from multiple viewpoints. Camera poses and dense scene geometry are reconstructed offline using COLMAP’s structure-from-motion (SfM) and multi-view stereo (MVS) pipelines~\cite{schoenberger2016sfm}. The resulting reconstruction is then scaled to metric units and aligned with the robot’s coordinate frame through fiducial-based registration.

\subsection{Viewpoint Sampling and Novel View Synthesis}
\label{sec:viewpoint_sampling_and_synthesize}
Using the metric-semantic map, we identify feasible mountable regions where our \ourmethod{} platform can be physically attached.
Given the vacuum-based mounting mechanism, we evaluate mounting feasibility using four criteria:
(i) geometric flatness sufficient for stable attachment,
(ii) sufficient area to accommodate the sensor module,
(iii) semantically inferred rigid, non-deformable structures that are compatible with suction mounting (e.g., walls, glass), while deformable objects such as curtains or soft fabrics are excluded, and
(iv) reachability by the robot during deployment.
Regions that satisfy all criteria are retained as candidate mountable regions.

To improve sampling efficiency, we bias viewpoint generation toward task-relevant regions.
We query a visually-conditioned language model (VLM) to identify task-relevant objects and preferentially sample viewpoints on mountable regions near these objects. For example, sampling is biased toward surfaces near the cutting board for a cutting task. We sample a set of candidate viewpoints:
\[
\mathcal{V} = \{ v_1, v_2, \ldots, v_N \},
\]
where each \( v_i \in SE(3) \) denotes a candidate viewpoint in the environment. 
For each candidate viewpoint \( v_i \), we synthesize a predicted
observation
$o_i = \mathcal{R}(v_i)$,
where \( \mathcal{R} \) denotes the radiance-field model for the environment \( E \). The rendered observation is later used to evaluate the task-conditioned utility of the corresponding viewpoint. Sampling proceeds coarse-to-fine over three rounds. The first round spreads 100 candidates over the feasible regions; each subsequent round samples 50 new candidates within 20\,cm and 30\textdegree{} of the 10 highest-scoring viewpoints from the previous round.

\subsection{Viewpoint Scoring and Selection}

The goal of viewpoint scoring is to select the most informative views for a given task.
Given the  predicted observation \(o\) and a task specification \(\tau\) (e.g., monitoring whether a human requires emergency assistance), we compute the utility:
$U(o \mid \tau)$,
which estimates the value of placing a sensor at that viewpoint for the task.  Different tasks may induce different utility functions, reflecting distinct perceptual objectives. For example, when observing a pot, a boil-over monitoring task prioritizes clear visibility of the pot rim and surface to detect overflow, whereas a meal preparation task may prioritize a wider view that captures the burner and surrounding utensils. We consider two task classes:
(1) \emph{object-coupled perception}, which prioritizes viewpoints that maximize information about detecting object state, and
(2) \emph{policy-coupled perception}, which prioritizes viewpoints compatible with a policy’s training distribution. 

We emphasize that $U(o \mid \tau)$ is a general abstraction: any function that ranks a given observation against a task specification can serve as a utility. The two instantiations below are concrete realizations we use in our experiments, not prescriptive choices. Alternative scoring mechanisms can be substituted without changing the overall framework.
\subsubsection{Object-Coupled Perception}

For object-coupled perception, the objective is to select the viewpoint that provides the most informative observation for detecting object state. As one realization, we instantiate the object-coupled utility:
\[
U_{\text{object}}(o \mid \tau) = U_{VLM}(o, \tau),
\]
which measures how informative the observation \(o\) is for distinguishing object states relevant to the task. 
Following prior work~\cite{ning2025prompting}, we query the VLM with candidate observations to evaluate which observation provides the most relevant information for the task \(\tau\).

\subsubsection{Policy-Coupled Perception}

For policy-coupled perception, the objective is to select viewpoints whose visual statistics align with the policy’s training distribution.
We instantiate the policy-coupled utility:
\[
U_{\text{policy}}(o \mid \tau)
=
\max_{e \in \mathcal{E}_{\text{train}\_\tau}}
\frac{e_{o}^\top e}{\|e_{o}\| \, \|e\|},
\]

where \(e_{o}\) denotes the feature embedding of a rendered observation \(o\), and
\(\mathcal{E}_{\text{train}\_\tau}\) is the set of embeddings extracted from training observations associated with task \(\tau\).

We treat training viewpoints as oracle viewpoints under which the policy performs reliably. We approximate these oracle viewpoints at deployment by selecting the candidate viewpoint whose embedding is most similar to the training embeddings. To construct \(\mathcal{E}_{\text{train}\_\tau}\), we use a pre-trained vision foundation model encoder (DINOv2~\cite{oquab2023dinov2}) to extract embeddings (grid of patch tokens, each L2-normalized and concatenated) from observations in the task-specific training dataset.
When the policy is fine-tuned on data collected for task \(\tau\), embeddings are computed from those fine-tuning observations, as they define the visual distribution under which the policy achieves reliable performance on task $\tau$.

During deployment, we compute an embedding \(e_{o}\) for each candidate rendered observation and evaluate its cosine similarity against all embeddings in \(\mathcal{E}_{\text{train}\_\tau}\).
The maximum similarity defines \(U_{\text{policy}}(o \mid \tau)\), and the viewpoint with the highest utility is selected for sensor placement.

\subsection{Sensor Detachment and Reattachment}
Once the highest-utility viewpoint is selected, we estimate the pose of the reconfigurable sensor platform. 
We perform model-based pose estimation using a CAD model of the magnetic suction stand (Fig.~\ref{fig:hardwaremerge}b) together with RGB-D observations from the wrist-mounted camera. We first obtain an initial alignment between the observed point cloud and the CAD model using geometric registration, and then refine the pose using colored iterative closest point (colored ICP), which jointly optimizes geometric and photometric consistency to improve alignment under real-world sensing noise. Given the refined pose estimate, the robot executes a grasp-and-place primitive to place the sensor platform at the selected viewpoint. The same pipeline is used for detachment and reattachment as task requirements change.

\section{Experiments}
\label{sec:Experiments}
We evaluate \ourmethod{} along three aspects:  (i) platform evaluation, measuring placement accuracy, attachment robustness, and continuous long-term sensing in real-world settings; (ii) task evaluation, validating that the system benefits downstream task outcomes; and (iii) system evaluation, quantifying computational cost to assess practical usability.

\begin{figure}[t]
\vspace{2mm}
\centering
\includegraphics[width=0.95\columnwidth]{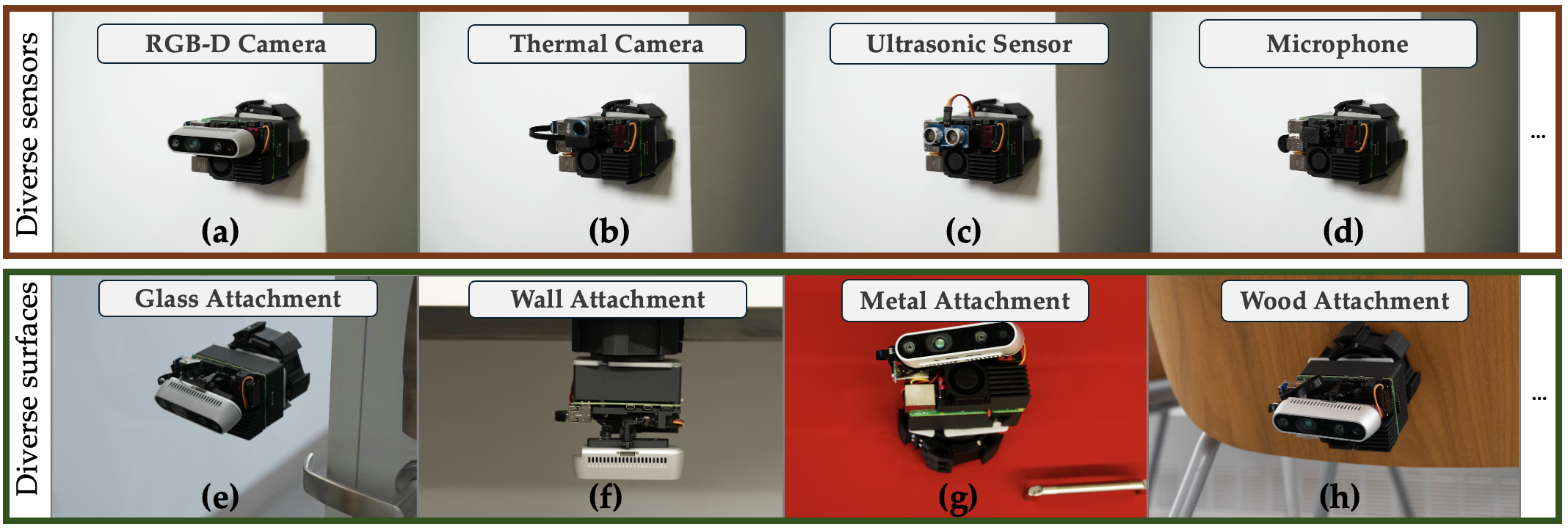}
\caption{
\textbf{Hardware versatility of \ourmethod{}.}
(a–d) Diverse sensing modalities supported by our platform.
(e–h) Reliable attachment across diverse surface materials.
}
\label{fig:sensor_versatility}
\vspace{-12pt}
\end{figure}

\subsection{Platform Evaluation}
\begin{figure}[t]
\centering
\includegraphics[width=\columnwidth]{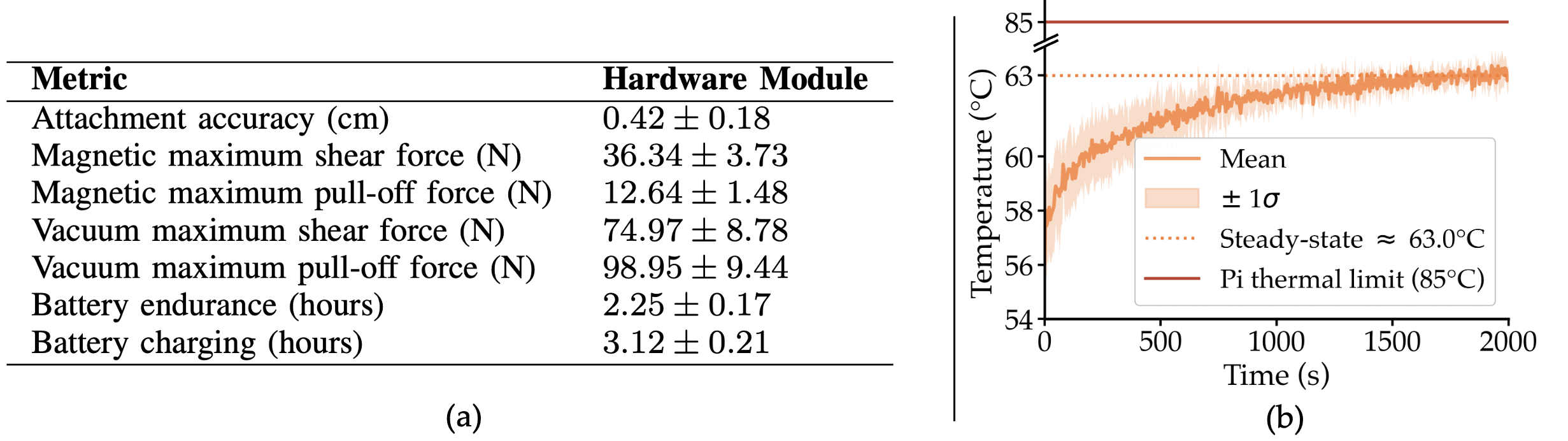}
\caption{
\textbf{Hardware performance evaluation of the \ourmethod{} platform.} (a) Hardware-level reliability metrics of the \ourmethod{}. Battery endurance is measured during continuous streaming with a RealSense D435i.
Values report mean $\pm$ standard deviation across 20 trials.
(b) Mean onboard temperature over time during sustained streaming, with $\pm1\sigma$ (standard deviation) shading. 
}
\label{fig:hardware_eval_latency_temp}
\vspace{-12pt}
\end{figure}

We develop the \ourmethod{} platform to support flexible viewpoint construction for real-world robot tasks. This requires our platform to have accurate placement, robust adhesion under disturbance, and sufficient onboard power. 
As shown in Fig.~\ref{fig:hardware_eval_latency_temp}a, attachment accuracy is $0.42 \pm 0.18$~cm, corresponding to sub-centimeter deviation and $<1\%$ positional error at typical workspace distances (80~cm). When sealed to a surface, our vacuum stand and magnetic interface between the stand and the sensor module provide strong mechanical support in both shear and pull-off directions (see Fig.~\ref{fig:hardware_eval_latency_temp}a). Given the total platform weight of 298.7~g ($\approx 2.92$~N) with the RealSense D435i payload, \ourmethod{} provides over $12\times$ weight resistance in shear and $4\times$ in pull-off, offering a substantial safety margin against disturbances.


Our onboard battery supports $2.25 \pm 0.17$ hours of continuous streaming with the RealSense D435i, exceeding the duration of typical household manipulation tasks. Our charging dock also enables automatic recharging upon return. Thermal measurements (Fig.~\ref{fig:hardware_eval_latency_temp}b) remain well below the platform’s throttling threshold, preventing performance degradation during extended use. 
Together, these results demonstrate that the \ourmethod{} hardware platform is capable of supporting long-horizon operation in the real world.


\begin{figure*}[ht]
\vspace{2mm}
\centering
\includegraphics[width=\textwidth]{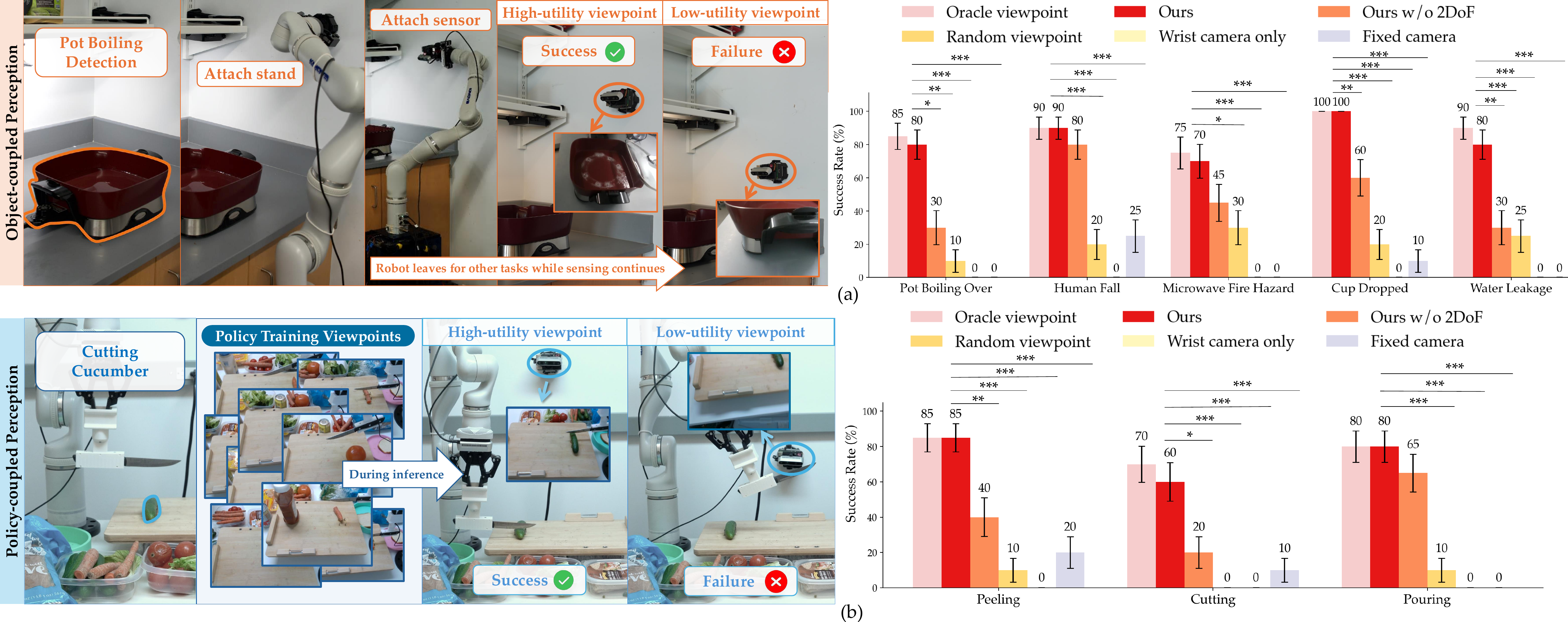}
\caption{
\textbf{Real-world demonstrations.}
\textbf{(a) Object-coupled perception.}
Left: The robot deploys the sensor to monitor pot state, enabling persistent event detection even when the robot leaves for other tasks.
Right: Quantitative comparison of event detection success rates over 20 trials.
\textbf{(b) Policy-coupled perception.}
Left: The robot autonomously deploys the \ourmethod{} platform to construct a third-person viewpoint for policy execution. 
Right: Quantitative comparison of task success rates over 20 trials. 
Significance: two-sided Fisher's exact test; * ($p < 0.05$), ** ($p < 0.01$) and *** ($p < 0.001$).}

\label{fig:example_demo_realworld}
\vspace{-12pt}
\end{figure*}

\subsection{Task Evaluation}
For task evaluation, we use a Kinova Gen3 7-DoF robotic arm equipped with a wrist-mounted Intel RealSense D435i RGB-D camera for egocentric perception, along with the \ourmethod{} platforms, each equipped with a D435i RGB-D camera for third-person viewpoint construction.

\textbf{Evaluation Scenarios.}
We evaluate \ourmethod{} on both object-coupled and policy-coupled perception tasks.
For \emph{object-coupled perception}, we test five real-world scenarios in which task-relevant events require persistent third-person observation for reliable object-state estimation after the robot leaves the workspace. 
For \emph{policy-coupled perception}, we evaluate three manipulation tasks in long-horizon meal preparation. For each task, we fine-tune a vision–language–action (VLA) $\pi_{0.5}$~\cite{intelligence2025pi05visionlanguageactionmodelopenworld} on 300 demonstrations collected with both wrist-mounted egocentric camera and a third-person camera. At test time, the third-person camera is removed while the wrist camera remains active. We evaluate whether \ourmethod{} can autonomously reconstruct the required third-person viewpoint to enable successful policy execution.

\textbf{Baselines.}
We compare SensorPerch against several baselines:
\textbf{(i) \ourmethod{} w/o 2-DoF adjustment.} The sensor platform is deployed without gimbal actuation, such that it remains perpendicular to the attached surface, isolating the contribution of viewpoint adaptability.
\textbf{(ii) Random feasible placement.} The sensor is attached to randomly selected mountable regions without task-aware viewpoint reasoning.
\textbf{(iii) Oracle viewpoint with manual reconfiguration.} For each task, a human manually repositions the sensor to a task-specific optimal viewpoint (e.g. the viewpoint on which the policy was trained), serving as an upper-bound reference for perception performance.
\textbf{(iv) Wrist camera only.} We only use the wrist-mounted camera, without any third-person viewpoint.
\textbf{(v) Fixed third-person camera.} A third-person camera is fixed to observe the whole environment and remains static throughout task execution.

\label{sec:Results}
\textbf{Object-coupled Perception Results.} We evaluate \ourmethod{} across five scenarios that require monitoring an object's state after the robot leaves the workspace: pot boiling over, human fall, microwave fire hazard, cup dropped, and water leakage. Across all five tasks (Fig.~\ref{fig:example_demo_realworld}a), \ourmethod{} achieves consistently high success rates, with no statistically significant difference from the oracle viewpoint ($p > 0.05$ across all tasks). It significantly outperforms random feasible placement ($p < 0.05$), wrist-camera-only sensing ($p < 0.001$), and fixed third-person sensing ($p < 0.001$) across all tasks. Removing the 2-DoF gimbal significantly degrades performance on pot boiling, cup dropped, and water leakage ($p < 0.01$), since a wall-mounted sensor must tilt downward to observe the target object. Random feasible placement reaches only 20\% success, showing that physical feasibility alone is insufficient without task-aware reasoning. The wrist camera fails once the robot departs, and a fixed third-person camera underperforms under occlusion and shifting viewpoint demands. Together, these results show that neither embodied sensing nor static infrastructure supports reliable long-horizon monitoring, and that task-conditioned sensor reconfiguration is necessary for robust object-state estimation.

\textbf{Policy-coupled Perception Results.}
Policy-coupled perception imposes a stricter requirement: the constructed view must not only be informative, but also must match the visual distribution the policy was trained on. We evaluate this on three daily tasks with distinct viewpoint needs: peeling, cutting, and pouring. \ourmethod{} achieves success rates of ${\sim}85$\%, ${\sim}80$\%, and ${\sim}60$\% on peeling, pouring, and cutting, respectively, with no statistically significant differences compared to the oracle viewpoint ($p > 0.05$ across all tasks), suggesting that the autonomously constructed views align with the policy’s training distribution. Removing 2-DoF adjustment significantly reduces success to ${\sim}40$\% for peeling ($p = 0.008$) and ${\sim}20$\% for cutting  ($p = 0.022$), while random placement fails almost entirely. These results show that policy performance is highly sensitive to viewpoint and that physical feasibility alone is insufficient. The wrist camera fails from self-occlusion, as the gripper and held tool (e.g., knife, peeler, bottle) block the interaction region. A fixed third-person camera also underperforms, since the optimal view differs across tasks (top-down for cutting, lateral for pouring), so no single static view serves all tasks. Together, these results show that policy success depends on reconstructing task-consistent viewpoints, which \ourmethod{} enables through reconfigurable sensing.

\begin{figure}[t]
  \centering
    \includegraphics[width=\linewidth]{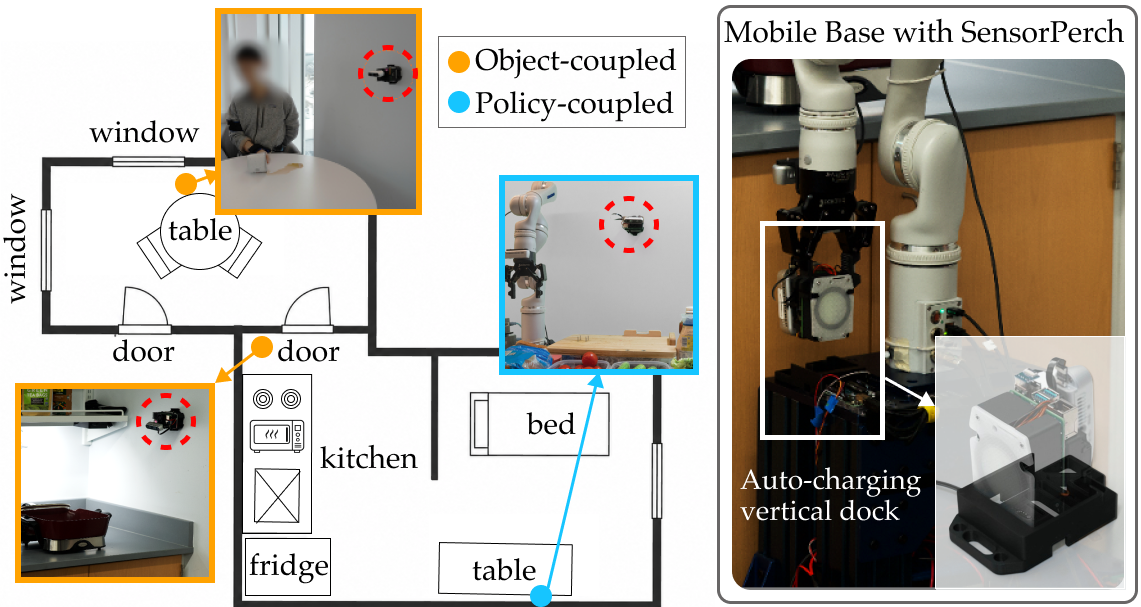}
    \caption{\textbf{Demonstration with multiple \ourmethod{} platforms in the real world}. 
    When a \ourmethod{} platform is idle, it is returned to the vertical docking station on the robot’s mobile base for automatic recharging.
    }
    \label{fig:multiple_sensor}
    \vspace{-12pt}
\end{figure}

\textbf{Real-world Demonstration with Multiple \ourmethod{} Platforms.}
We demonstrate multiple \ourmethod{} platforms in a laboratory environment simulating a household (Fig.~\ref{fig:multiple_sensor}). While preparing breakfast and assisting a human, the robot deploys three platforms, each serving a distinct sensing role: monitoring boiling water, providing a task-relevant viewpoint for cucumber cutting, and monitoring the human's coffee cup. Upon detecting boiling water or an empty cup, the robot pauses cutting to turn off the stove or refill the cup, then resumes cutting. Across 10 trials, the system achieves a 90\% success rate in detecting task-relevant object states and an 80\% success rate in cutting policy execution. When no longer needed, platforms are returned to a vertical docking station on the robot's mobile base for automatic recharging, ensuring charged platforms are available for future sensing tasks.

\begin{figure}[t]
\vspace{2mm}
\centering
\includegraphics[width=\columnwidth]{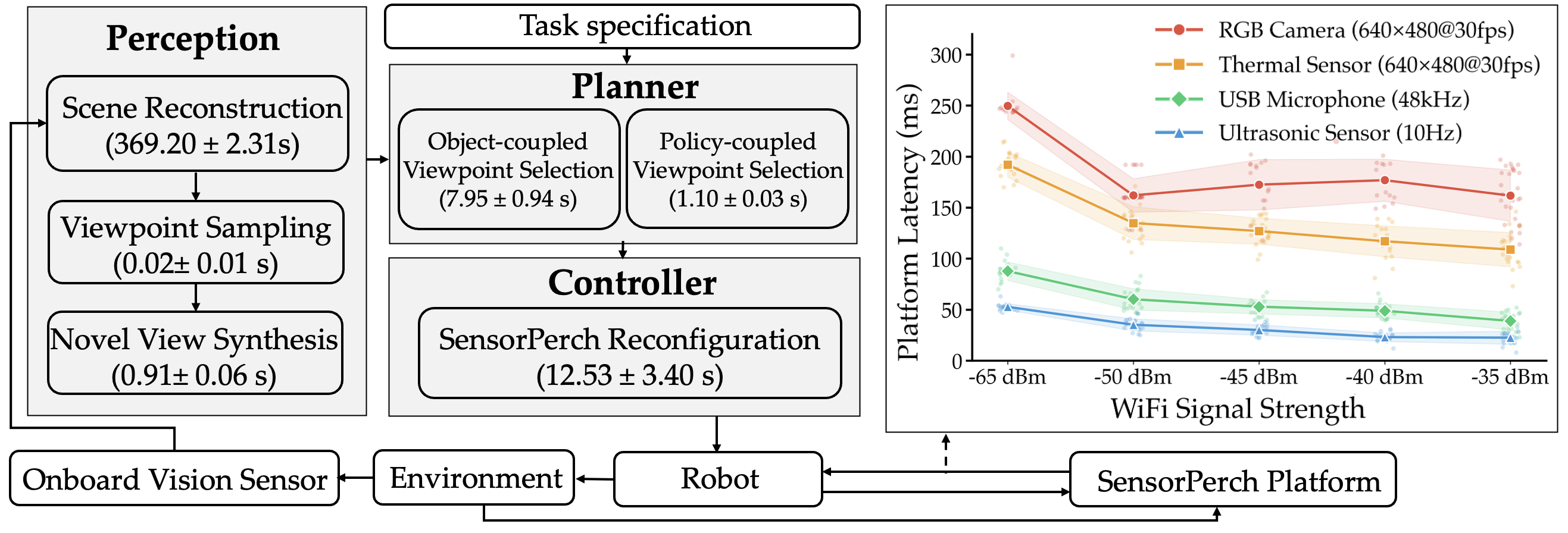}
\caption{
\textbf{System latency.} We report the latency of each component in our system, and specifically the SensorPerch platform streaming latency under different Wi-Fi signal strengths across multiple sensing modalities. Results are reported as mean $\pm$ standard deviation across 20 trials.
}
\label{fig:system_latency}
\vspace{-12pt}
\end{figure}

\subsection{System Evaluation}

Beyond task effectiveness, we analyze the computational efficiency of our system to assess practical usability. Fig.~\ref{fig:system_latency} reports the latency breakdown of the system and the streaming latency of the SensorPerch platform. We first report the latency of each component. Scene reconstruction requires $369.20 \pm 2.31$ seconds and is only performed once per environment. 
This radiance-field scene reconstruction is \textbf{optional}. Alternative approaches for scene reconstruction or viewpoint selection can also be incorporated into our framework.
Viewpoint sampling and synthesis require $0.91 \pm 0.05$ seconds per round. Viewpoint scoring takes $7.95 \pm 0.94$ seconds for object-coupled perception and $1.10 \pm 0.03$ seconds for policy-coupled perception. The higher latency in object-coupled scoring stems from parallel VLM queries, whereas policy-coupled scoring relies on precomputed feature embeddings. 
\ourmethod{} platform reconfiguration (the complete detach-reattach loop) requires $\mathbf{12.53 \pm 3.40}$ seconds, which is small relative to task execution time and avoids manual reconfiguration between tasks. Otherwise, the robot would have to rely on human intervention to reconfigure the sensor before proceeding to the next task.
We next evaluate streaming reliability. With a RealSense D435i camera, the platform achieves an average streaming latency of approximately $180$~ms and remains below $250$~ms even under weak Wi-Fi signal strength ($-65$~dBm, $\sim$15 meters from the router with a 433.3~Mbit/s link rate), which supports real-time perception. We also test multiple sensing modalities and run three sensor platforms simultaneously on the same network, observing no measurable increase in streaming latency compared to a single module.
These results demonstrate that \ourmethod{} constructs and sustains task-adaptive viewpoints at latencies compatible with real-world operation.

\section{Discussion and Limitations}
\label{sec:Discussions}

\ourmethod{} treats robot perception as a problem of physically constructing task-relevant viewpoints, rather than constraining sensing to the robot’s embodiment or fixed infrastructure. For long-horizon tasks, our platform can be detached and re-attached between task stages if requirements change. By modeling sensors as independent physical entities, we enable perception to adapt to task demands. However, this capability depends on the availability of attachable surfaces and the reliability of vacuum-based mounting, which may be challenged by porous or compliant surface materials. Incorporating more explicit surface material estimation could further enhance deployment robustness. 
Finally, the current framework does not explicitly reason about how robot motion may create or remove occlusions. Explicitly modeling how robot motion affects visibility could further improve perception robustness.

\bibliographystyle{IEEEtran}
\bibliography{references}


\end{document}